%% file: main.tex
\documentclass{article}
\usepackage{spconf,amsmath,graphicx}

\usepackage{microtype}
\usepackage{graphicx}
\usepackage{subcaption}
\usepackage{booktabs} % for professional tables
\usepackage{times}  %Required
\usepackage{helvet}  %Required
\usepackage{courier}  %Required
\usepackage{url}  %Required
\graphicspath{{figures/}}
\usepackage[utf8]{inputenc} % allow utf-8 input
\usepackage[T1]{fontenc}    % use 8-bit T1 fonts
\usepackage{hyperref}
\usepackage{amsfonts}       % blackboard math symbols
\usepackage{nicefrac}       % compact symbols for 1/2, etc.
\usepackage[flushleft]{threeparttable}
\usepackage{enumitem}
\usepackage{float}
\usepackage{epsfig}
\usepackage{amsmath,amsthm,amssymb}
\usepackage{multirow}
\usepackage[ruled,algo2e]{algorithm2e}
\usepackage{algorithm}
\usepackage{xcolor}
\usepackage{shadethm}
\usepackage{thmtools}
\declaretheoremstyle[%
  spaceabove=-6pt,%
  spacebelow=6pt,%
  headfont=\bfseries\itshape,%
  postheadspace=0.5em,%
  qed=\qedsymbol%
]{mystyle} 

\declaretheorem[
  shaded={rulecolor=black, rulewidth=1pt, bgcolor=blue!5!white},
  name=Theorem,
]{theo}

\declaretheorem[name={Proof},style=mystyle,unnumbered,
]{Proof}
\theoremstyle{mystyle}

\newtheorem{assumption}{Assumption}
\newtheorem{corollary}{Corollary}
\newtheorem{lem}{Lemma}
\newtheorem{proposition}{Proposition}
\newtheorem{defn}{Definition}
\newtheorem{rem}{Remark}

\newcommand{\x}{\mathbf{x}}

\newcommand{\X}{\mathbf{x}}

\DeclareMathOperator*{\argmin}{arg\,min}
\title{Decentralized Deep Learning using Momentum-Accelerated Consensus}
\name{\small Aditya Balu$^{\star}$ $\qquad$ Zhanhong Jiang$^{\dagger}$ $\qquad$ Sin Yong Tan$^{\star}$ $\qquad$ Chinmay Hedge $^{\S}$ $\qquad$ Young M Lee$^{\dagger}$ $\qquad$ Soumik Sarkar$^{\star}$\thanks{corr address: soumiks@iastate.edu}}
  
\address{\small $^{\star}$ Iowa State University $\qquad$
      $^{\dagger}$ Johnson Controls $\qquad$
      $^{\S}$ New York University}

\begin{document}
\ninept
\maketitle
% \newcommand{\nnfootnote}{
%   \begin{NoHyper}
%   \printAffiliationsAndNotice{\icmlEqualContribution} % otherwise use the standard text.
%   \end{NoHyper}
% }
% \nnfootnote{}

\begin{abstract}
We consider the problem of decentralized deep learning where multiple agents collaborate to learn from a distributed dataset. While several decentralized deep learning approaches exist, the majority consider a central parameter-server topology for aggregating the model parameters from the agents. However, such a topology may be inapplicable in networked systems such as ad-hoc mobile networks, field robotics, and power network systems where direct communication with the central parameter server may be inefficient. In this context, we propose and analyze a novel decentralized deep learning algorithm where the agents interact over a fixed communication topology (without a central server). Our algorithm is based on the heavy-ball acceleration method used in gradient-based optimization. We propose a novel consensus protocol where each agent shares with its neighbors its model parameters and gradient-momentum values during the optimization process. We consider nonconvex objective functions and theoretically analyze our algorithm's performance. We present several empirical comparisons with competing decentralized learning methods to demonstrate the efficacy of our approach under different communication topologies. 

\end{abstract}

\section{Introduction}
Spurred by the need to accelerate deep neural network training with massive distributed datasets, several recent research efforts~\cite{dean2012large,zhang2015deep,jin2016scale,kairouz2019advances} have put forth a variety of distributed, parallel learning approaches. One line of work has focused on adapting traditional deep learning algorithms that use a single CPU-GPU environment to a distributed setting with a network of several GPUs~\cite{wen2017terngrad,zhang2015deep,goyal2017accurate,chen2016revisiting}. Some of these approaches also can be used in conjunction with gradient compression schemes between compute nodes in the network~\cite{bernstein2018signsgd}. A different line of works falls under the umbrella of \emph{federated learning}~\cite{konevcny2016federated} which deals with inherently decentralized datasets, i.e., each compute node has its own corresponding set of data samples that are not shared. The majority of works in this area consider a central parameter-server topology that aggregates estimates of model parameters from the agents. 

In this paper, our particular focus is on decentralized learning where there is \emph{no} central server: each node in the network maintains its model parameters (which it can communicate with its neighbors defined according to a pre-specified, but otherwise arbitrary, communication topology), and the goal is to arrive at a consensus model for the whole network. See~\cite{lian2017can,jiang2017collaborative,assran2018stochastic,kamp2018efficient} for examples of such decentralized learning approaches.

\setlength\tabcolsep{2pt}
\begin{table*}[!t]
\begin{center}
\caption{Comparisons between different optimization approaches. 
Gra.Lip.: Gradient Lipschitz. Str.Con.: strongly convex. Cen.: centralized. Con.: convex. Dec: decentralized. $\rho$: a positive constant in $(0,1)$. $k$ is the number of iterations. $N$: the number of agents. PL: Polyak-\L{}ojasiewicz condition. It should be noted that each $\rho$ in different methods vary in real values.
}
\label{table1}
  \begin{tabular}{c c c c c c c}
    \hline
    Method & $f$ & Rate & Setting & Gra.Lip. & Stochastic&Momentum\\ \hline
    HBM & Str.Con. & $\mathcal{O}(\rho^k)$ & Cen. &Yes & No&Yes\\
    MSWG~\cite{ghadimi2013multi} & Str.Con. & $\mathcal{O}(\rho^k)$ & Dec.& Yes &No&Yes\\
    SHB~\cite{loizou2017linearly} & Con. & $\mathcal{O}(\rho^k)$ & Cen. &Yes & Yes&Yes \\
    \textbf{DMSGD (This paper)} & \textbf{PL (Quasi-convex)} & $\mathcal{O}(\rho^k)$  & \textbf{Dec.} & \textbf{Yes} & \textbf{Yes}&\textbf{Yes}\\ \hline
    SUM~\cite{yang2016unified} & Nonconvex & $\mathcal{O}(1/\sqrt{k})$ & Cen. &Yes &Yes&Yes\\
    CDSGD/D-PSGD & Nonconvex & $\mathcal{O}(1/k+1/\sqrt{Nk})$ & Dec.& Yes &Yes&No \\
    MSGD~\cite{yu2019linear} & Nonconvex&$\mathcal{O}(1/k+1/\sqrt{Nk})$ & Cen./Dec.& Yes&Yes&Yes\\
    SlowMo\cite{wang2019slowmo} & Nonconvex&$\mathcal{O}(1/k+1/\sqrt{Nk})$&Cen.&Yes&Yes&Yes\\
    \textbf{DMSGD (This paper)} & \textbf{Nonconvex} & $\mathcal{O}(1/k+1/\sqrt{Nk})$ & \textbf{Dec.} & \textbf{Yes} &\textbf{Yes}&\textbf{Yes} \\
    \hline
  \end{tabular}
\vspace{-20pt}
\end{center}
\end{table*}

While the above works are representative of key advances in the algorithmic front, several gaps remain in our understanding of centralized versus distributed learning approaches. Conspicuous among these gaps is the notion of \emph{momentum}, which is a common technique to speed up convergence in gradient-based learning methods~\cite{nesterov2013introductory,sutskever2013importance}. However, few papers (barring exceptions such as~\cite{ghadimi2013multi,jiang2017collaborative,assran2018stochastic,yu2019linear}) in the decentralized learning literature have touched upon momentum-based acceleration techniques, and to our knowledge, rigorous guarantees in the context of nonconvex and stochastic optimization have not been presented. Our objective in this paper is to fill this key gap from both a theoretical as well as empirical perspective. 

% The standard analysis of decentralized learning algorithms establishes convergence rates of gradient-based learning assuming that the loss function is strongly convex. However, strong convexity is too strict an assumption in real-world applications, and particularly in the deep learning setting. In the centralized case, one way to overcome this is to assume objective functions that obey the Polyak-\L{}ojasiewicz, or the PL, criterion (also called as quasi-convex or invex functions). It has been shown that SGD and its variants algorithms achieve global linear convergence rates~\cite{karimi2016linear,gower2019sgd}. However, such analysis has not performed in the decentralized setting to our knowledge. In this paper, we first prove linear convergence rate of momentum-accelerated methods with constant stepsize for strongly convex functions. Then, we relax the strongly-convex assumption to the quasi-convex case and prove similar (linear) convergence rates.
 
\textbf{Our contributions.} 
We propose and analyze a stochastic optimization algorithm that we call decentralized momentum SGD  (DMSGD), based on the classical notion of momentum (or the \emph{heavy-ball} method~\cite{polyak1964some}). %In contrast with previous analyses~\cite{jakovetic2014fast,ghadimi2013multi}, our work takes into account the following factors: DMSGD succeeds in the \textit{stochastic} setting with \textit{decentralized} data/parameters for optimizing objective functions that exhibit \textit{quasi-convexity} (in the PL sense) as well as \textit{nonconvexity}.
See Table~\ref{table1} for more comparisons.

For smooth and nonconvex objective functions, we show the convergence to a \textit{first-order stationary point}, that is, the algorithm produces an estimate $x$ with sufficiently small gradient ($\|\nabla f(x)\|\leq \varepsilon$) after $\mathcal{O}(1/\varepsilon+1/(N\varepsilon^2))$ iterations, where $N$ is the number of agents. Additional results on strongly-convex and its relaxation using the Polyak-\L{}ojasiewicz criterion is provided in the supplementary materials. %For smooth and strongly-convex objective functions, when the step-size is appropriately chosen to be constant, we first establish that the convergence rate of the proposed algorithm is linear. We then relax the requirement of strong-convexity using the Polyak-\L{}ojasiewicz criterion and obtain the convergence rate to be linear in the quasi-convexity. In the main content, we only present the analysis for the nonconvex functions.

We empirically compare DMSGD with baseline decentralized methods such as D-PSGD/CDSGD~\cite{lian2017can,jiang2017collaborative}. We show that when the momentum term is appropriately weighted, DMSGD is faster and more accurate than these baseline methods, suggesting the benefits of its use on practice.

\vspace{-10pt}
\input{NewAnalysisMain.tex}
\begin{figure*}[!ht]
\centering
\begin{subfigure}[t]{0.325\linewidth}
\includegraphics[width=\linewidth,trim={3cm, 2.5cm, 3cm, 3cm},clip]{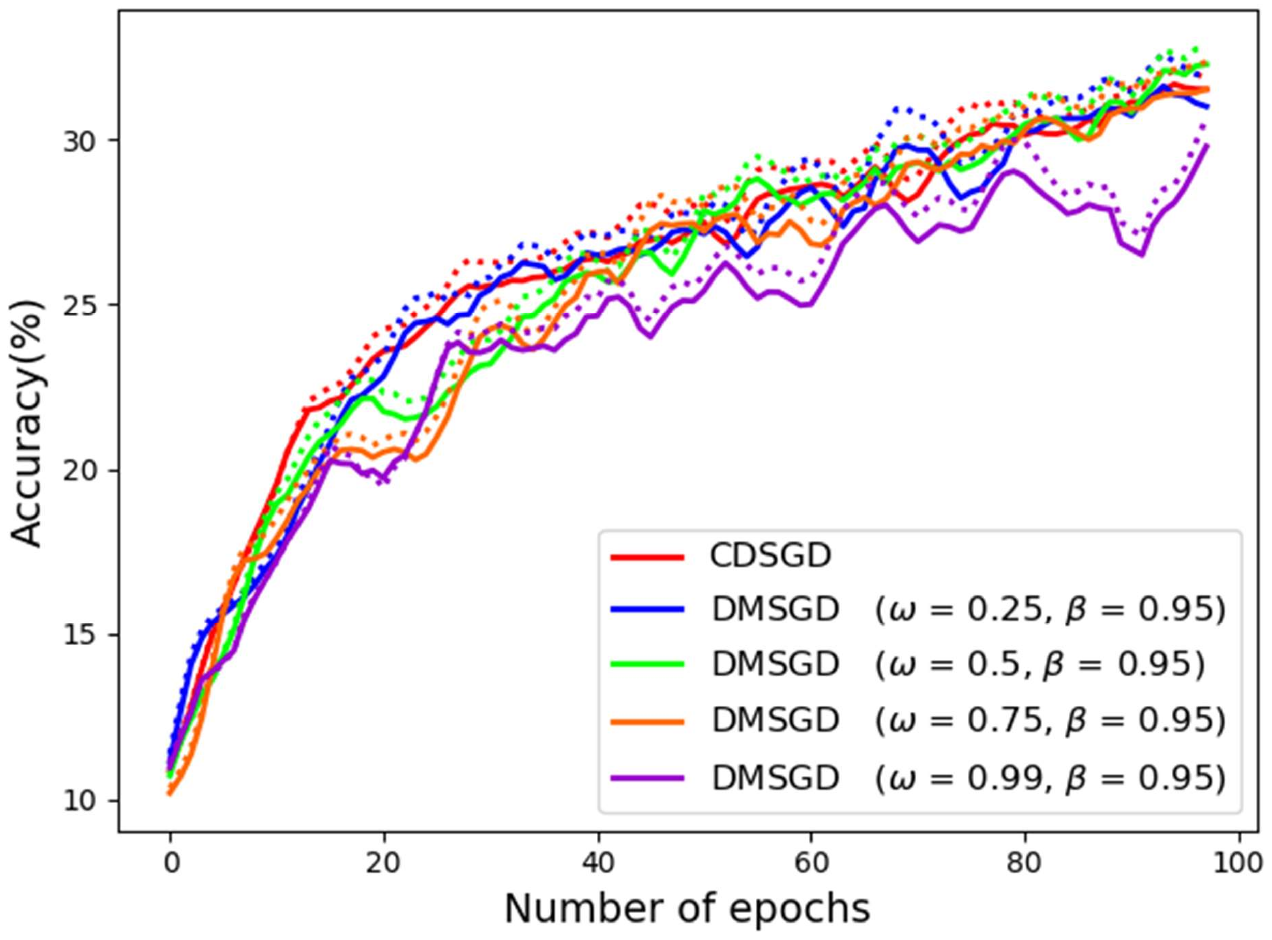}\vspace{-5pt}
\caption{}
\end{subfigure}
\begin{subfigure}[t]{0.325\linewidth}
\includegraphics[width=\linewidth, trim={3cm, 2.5cm, 3cm, 3cm},clip]{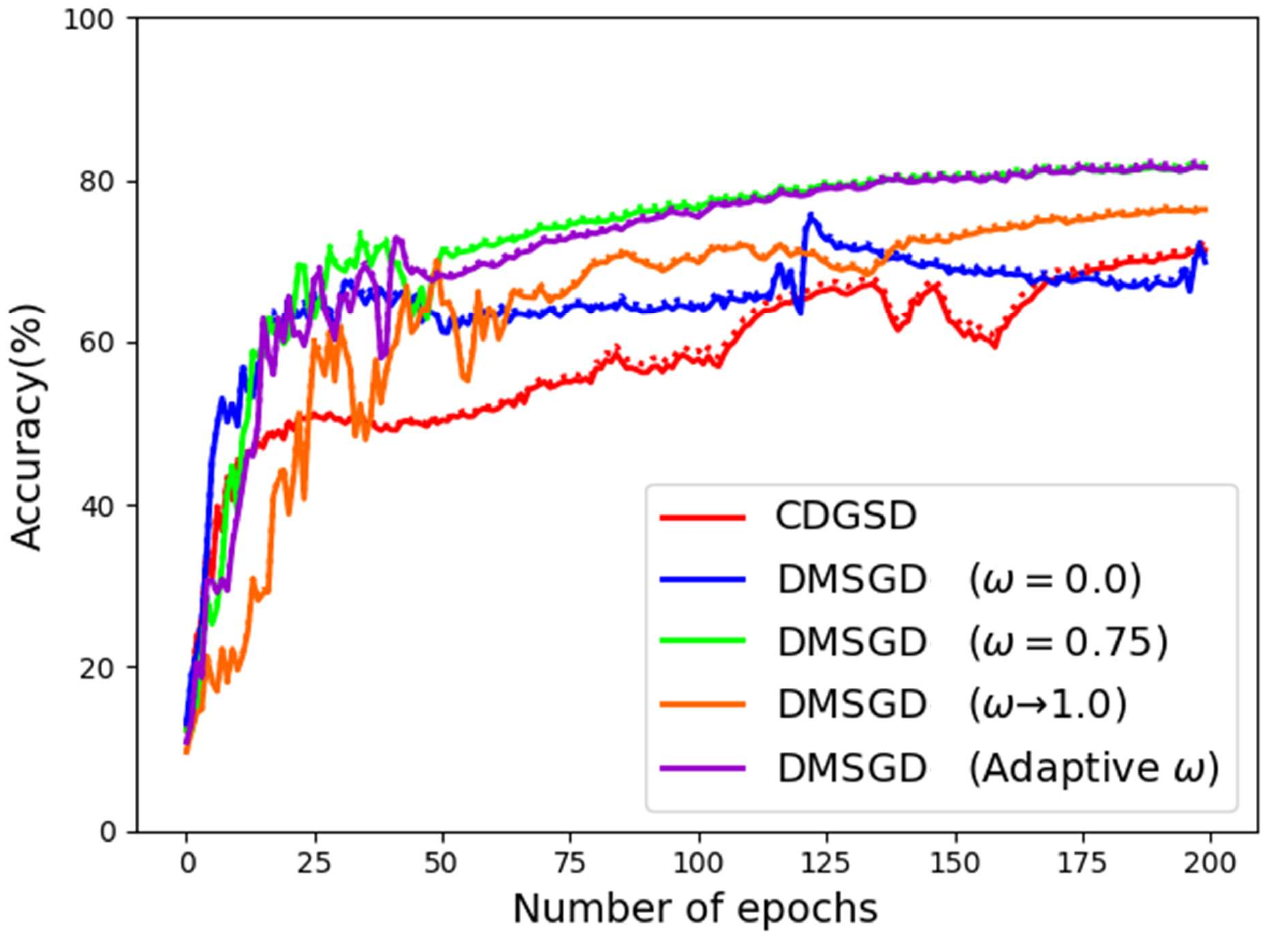}\vspace{-5pt}
\caption{}
\end{subfigure}
\begin{subfigure}[t]{0.325\linewidth}
\includegraphics[width=\linewidth, trim={0.05in 0.1in 0.15in 0.15in}, clip]{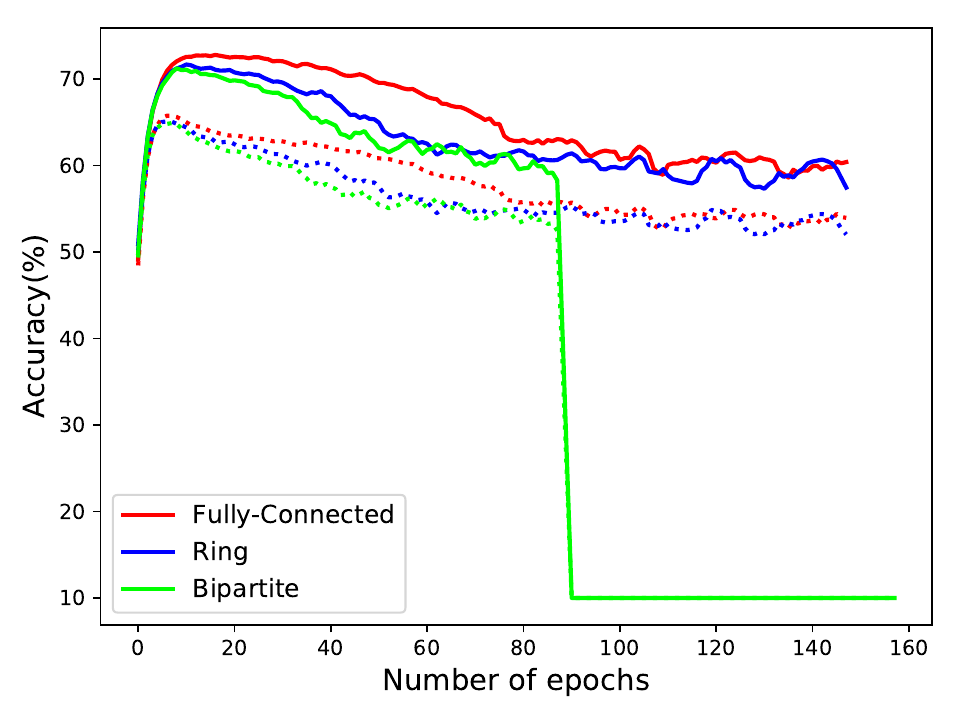}
\caption{}
\end{subfigure}
\vspace{-10pt}
\caption{(a)~Performance of our proposed algorithm, DMSGD with different $\omega$ values, and its comparison with CDSGD. These performances are with iid data simulation strategy (b) Performance of our proposed algorithm in non-iid data simulation strategy (c) Performance on different topologies for mnist dataset.}
\vspace{-12pt}
\label{Fig:2}
\end{figure*}

\section{Experimental Results}\label{expe}
We now support the utility of our proposed DMSGD algorithm by simulating a distributed environment over a GPU cluster with multiple GPUs, similar to the experiments of~\cite{lin2014accelerated,jiang2017collaborative,tang2018d2}. We define a graph topology where each agent in the graph can communicate with another agent with an interaction matrix initialized by the user, ensuring that it is doubly stochastic (in our experiments, we explore a fully connected topology, a ring topology, and a bipartite graph just as in~\cite{assran2018stochastic}). 

We split the given (complete) training dataset among different agents equally, creating two data simulation strategies:
\begin{enumerate} 
\item~\textbf{iid:} the dataset is shuffled completely and distributed amongst the agents to simulate an environment where each of the agents has an independently identical draw from the data distribution. 
\item~\textbf{non-iid:} We first segregate the dataset based on the target labels, then we create chunks of data and distribute the chunks with unique target labels to all the agents. If the number of agents is larger than the number of target labels, each agent gets only a chunk of data corresponding to each target label, and if the number of agents is lesser than the number of target labels, each agent gets a set of multiple chunks with unique target labels unavailable with other agents. This strategy simulates an extreme imbalance across different agents and we expect to see significant loss in the performance of  decentralized learning algorithms.
\end{enumerate}
In this work, we implement proposed algorithms with both the data simulation strategies. The implementation is using Pytorch~\cite{paszke2017pytorch}.

First, we demonstrate empirical evidence of good consensus using the lesser generalization gap as done by \cite{jiang2017collaborative,tang2018d2}. In Figure~\ref{Fig:2}, the dotted lines denote the performance of agents on test data, which closely follow the solid line (performance on training data) but lag slightly.  We attribute to the averaging of several weights, which promotes generalization, as explained in~\cite{izmailov2018averaging,huang2017snapshot}. In \cite{izmailov2018averaging}, authors show that by averaging the weights of the network, they get wider and flat optima that generalize well. We note that the consensus step provides us with similar conditions. Another observation from our experiments is a validation of Remark~\ref{rem:weakness}; we see that as $\omega$ increases, the generalization gap increases with a weaker consensus bound occurring at $\omega \to 1$ as explained in Remark~\ref{rem:weakness}. Therefore, we see that at $\omega=0.99$, our algorithm does not converge.

Now, we analyze the convergence and performance of the DMSGD algorithm. Due to space constraints, we only present a few anecdotal results here. In Figure~\ref{Fig:2}(a and b), we show the performance of DMSGD with different $\omega$ values for CIFAR-10 dataset. All the results shown here are for a sufficiently large Convolutional Neural Network. While we could perform comparisons with the algorithm proposed by~\cite{assran2018stochastic,kamp2018efficient}, it would be unfair as the protocol for communication used by them is different (Push-Sum and Dynamic Model Averaging). Note that we could extend our momentum-accelerated consensus to these models, analysis of the same is beyond the scope of this work. Therefore, as a baseline, we use a non-momentum decentralized algorithm that would have a fair comparison.  For this, we compare with CDSGD~\cite{jiang2017collaborative} in this simple experiment. We observe that DMSGD performs with similar performance as the CDSGD algorithm, i.e., without any acceleration. However, while working on a non-iid data simulation strategy, DMSGD performs better than the CDSGD algorithm. We believe that this is a trade-off between consensus and convergence, which~\cite{li2019communication} explores in detail.  

We also note from the results shown and the analysis in the previous section that as $\omega \to 1$, the convergence bounds become weaker. This explains why the performance dies down as a function of $\omega$, e.g. $\omega=0.5$ performs better than $\omega=0.75$. However, setting $\omega=0$ performs very badly for non-iid data. The dynamics of $\omega$ with respect to the data distribution is not explored in this work and can be considered as future work.

Finally, we would like to add another result for the performance of our proposed DMSGD algorithm for different communication topologies in Figure~\ref{Fig:2}(c). We consider three communication topologies: (1) Fully connected topology (2) Ring topology (3) Bipartite topology. As the communication topology has sparse communication, the consensus and convergence bounds also become weaker. In Figure~\ref{Fig:2}(c), where we see that the Bipartite graph with very sparse connections performs worse than fully connected graph, which validates the analysis.

% \begin{figure}[!ht]
%  \centering
% 	\includegraphics[width=0.8\linewidth, trim={0in 0.1in 0in 0in}, clip]{Figures/topo_cifar10_D1_beta_95_graph_avg_accs.pdf}
%     \caption{Performance on different topologies on DMSGD-I algorithm for mnist dataset.}
%   \label{Fig:3}
% \end{figure}

\section{Conclusions and Future Work} %\label{conclusions}
This paper addresses the problem of deep learning in a decentralized setting using momentum accelerated consensus. We establish a consensus-based decentralized learning algorithm using the stochastic heavy ball approach that can assist in finding the optimal solution faster than conventional SGD-style methods. We show that the proposed DMSGD with different choices of momentum terms can achieve linear convergence rate with appropriately chosen step size for strongly-convex, quasi-convex objective functions along with the assumption of smoothness, and convergence to a stationary point for nonconvex objective functions.

Relevant experimental results using benchmark datasets show that the proposed algorithms can achieve better accuracy with sufficient training epochs. While our current research focuses on extensive testing and validation of the proposed framework (especially for large networks), a few directions for future research include an extension to the analysis of Nesterov momentum with nonconvex objective functions, analysis of non-iid data setting and variance reduction strategies for further convergence speed-up techniques in the stochastic setting. 

\bibliographystyle{IEEEbib}
\bibliography{references}
\clearpage

% \appendix

% %%%%%%%%%%%%%%%%%%%%%%%%%%%%%%%%%%%%%%%%%%%%%%%%%%%%%%%%%%%%%%%%%%%%%%%%%%%%%%%
% %%%%%%%%%%%%%%%%%%%%%%%%%%%%%%%%%%%%%%%%%%%%%%%%%%%%%%%%%%%%%%%%%%%%%%%%%%%%%%%
\section*{Supplementary Materials for ``Decentralized Deep Learning using Momentum-Accelerated Consensus"}\label{supp}
% This section presents additional analytical and experimental results. The statements of the lemmas and theorems are presented with their detailed proof for completeness.

\input{NewAnalysisSupp.tex}
\end{document}

%% file: NewAnalysisMain.tex
\section{Problem Setup and Preliminaries}\label{algorithm}
Let the parameters of the deep neural network be denoted as $x\in \mathbb{R}^{d}$. We define a loss function $f:\mathbb{R}^d\to\mathbb{R}$ and denote its corresponding stochastic gradient by $g$.

\textbf{Decentralized learning}. Consider a static undirected graph $G = (V,E)$, where $V$ is the node set and $E$ is an edge set. Consequently, if we assume that there exist $N$ nodes (agents) in the networked system, we can denote $V = \{1,2,...,N\}$ while $E \subseteq V\times V$. If $(j,l)\in E$, then agent $j$ can communicate with agent $l$. A node $j\in V$ has its neighbors $Nb(j)\triangleq\{j\in V:(j,l)\in E\;\textnormal{or}\;l=j\}$. We assume that the network $G$ is {connected} without loss of generality throughout this paper. In this paper, we consider a finite sum minimization problem defined as follows:\vspace{-7pt}
\begin{equation}\label{eq1}
\vspace{-7pt}
\text{min}\; \frac{1}{n}\sum_{j=1}^{N}\sum_{i\in\mathcal{D}_j}f^i_j(\X) ,\\
\end{equation}
where $\mathcal{D}_j$ denotes the subset of the training data (comprising $n_j$ samples) only known by the $j^{\textrm{th}}$ agent such that $\sum_{j=1}^N n_j=n$, $n$ is the size of dataset, $N$ is the number of agents, $f_j: \mathbb{R}^d\to \mathbb{R}$ are local loss functions of each node. Let $x^j \in \mathbb{R}^d$ be a local copy of $x$.  Then, define $\X = [x^1; x^2; \dots; x^N] \in \mathbb{R}^{Nd\times 1}$. All vector and matrix norms are Euclidean and Frobenius norms respectively.

In this paper, for simplicity of presentation, we assume that $d=1$, while noting that exactly the same proof ideas hold when $d>1$ albeit at the expense of extra notation.

Equation~\ref{eq1} can be rewritten as the constrained problem:\vspace{-10pt}
\begin{equation}
\label{eq2}
%\begin{alignat}{2}\vspace{-10pt}
\vspace{-7pt}
\text{min}\;F(\x)\triangleq\frac{1}{n}\sum_{j=1}^{N}\sum_{i\in\mathcal{D}_j}f^i_j(x^j),~~~\textnormal{s.t.}~~\Pi\x=\x,
%\end{alignat}
\end{equation}
where the matrix $\Pi$ is the mixing matrix encoding the adjacency structure of $G$ (which is assumed to be \textit{doubly stochastic}). By turning the hard constraint $\Pi\x=\x$ into a soft constraint that penalizes the corresponding decision variables $\x$, the following equivalent objective function can be obtained:
\begin{equation}\label{unified_objective_1}\vspace{-5pt}
    \mathcal{F}(\x):=F(\x)+\frac{1}{2\xi}(\x^T(I-\Pi)\x)
\end{equation}
where $\xi> 0$. In the next section, we will show that $\xi$ can be related to the step size $\alpha$.
%We use $\mathcal{F}(\x)$ for showing the main results.

% \paragraph{\DMSGD~Algorithm:}
In order to study the behavior of the proposed  algorithm, we now present basic definitions and assumptions.

\begin{defn}\label{smooth}
A function $f:\mathbb{R}^d\to \mathbb{R}$ is $L$-smooth, if for all $x,y\in \mathbb{R}^d$, we have
$$f(y)\leq f(x)+\nabla f(x)^T(y-x)+\frac{L}{2}\|y-x\|^2.$$
\end{defn}

\begin{defn}\label{coercivity}
A function $c(\cdot)$ is said to be coercive if it satisfies
$c(x)\to \infty\;when\;\|x\|\to \infty.$
\end{defn}

\begin{assumption}\label{assump_objective}
The objective functions $f_j:\mathbb{R}^d\to\mathbb{R}$ are assumed to satisfy the following conditions: a) Each $f_j$ is $L_j$-smooth; b) each $f_j$ is proper (not everywhere infinite) and coercive.
\vspace{-10pt}
\end{assumption}
An immediate consequence of Assumption~\ref{assump_objective} a) is that $\sum_{j=1}^{N}f_j(x^j)$ is $L_m$-smooth where $L_m:=\textnormal{max}\{L_1, L_2, ..., L_N\}$. 
\begin{assumption}\label{assump_lip}
The unified objective function $\mathcal{F}(\X)$ has bounded gradient such that $\|\nabla\mathcal{F}(\mathbf{x})\|\leq M$.
\end{assumption}
Denote $\mathcal{S}(\x)$ by the stochastic gradient of $\mathcal{F}$ at point $\x$ such that it is the unbiased estimate of $\nabla\mathcal{F}(\mathbf{x})$. We next make another assumption on the variance of $\mathcal{S}(\X)$ to ensure that it is bounded from above.
\begin{assumption}\label{assump_var}
The stochastic gradients of $\mathcal{F}$ satisfy: 
% \begin{align}\nonumber
$Var(\mathcal{S}(\X)) = \mathbb{E}[||\mathcal{S}(\X) - \nabla \mathcal{F}(\X)||^2] \leq \sigma^2$\\
% (b) &~~\nabla \mathcal{F}(\X) = \mathbb{E}[\mathcal{S}(\X)]\nonumber.
% \end{align}\vspace{-30pt}
\end{assumption}
% Assumption~\ref{assump_var}(b) implies that $\mathcal{S}(\X)$ is an unbiased estimate of $\nabla \mathcal{F}(\X)$. The above assumptions also imply the boundedness of the stochastic gradient of $\mathcal{F}(\X)$. As $\mathcal{F}(\X)$ is smooth, $||\nabla \mathcal{F}(\X)||$ is also bounded above by $M$. 
With Assumption~\ref{assump_lip} and Assumption~\ref{assump_var}(b), we have \vspace{-10pt}
\begin{align}\nonumber
    \mathbb{E}[||\mathcal{S}(\X)||] &= \sqrt{(\mathbb{E}[||\mathcal{S}(\X)||])^2} \leq \sqrt{\mathbb{E}[||\mathcal{S}(\X)||^2]}\\\nonumber
                        &= \sqrt{||\mathbb{E}[\mathcal{S}(\X)]||^2 + Var(S(\X))}\\\nonumber
                        &\leq \sqrt{M^2 + \sigma^2}.\vspace{-20pt}
\end{align}\vspace{-10pt}
\vspace{-20pt}
\section{Proposed Algorithm}
We first present our proposed approach in Algorithm~\ref{dmsgd_algorithm}.
\setlength{\textfloatsep}{0pt}
\begin{algorithm}[!t]
    \caption{DMSGD\label{dmsgd_algorithm}}
    \SetKwInOut{Input}{Input}
    \SetKwInOut{Output}{Output}

    \Input{$m$, $\Pi$, $x^j_0, x^j_1$, $\alpha$, $N$, $\omega$, $\beta$}
    \Output{$x^*$}
    \For{$k=1:m$}
      {
        \For{$j=1:N$}
          {
            \text{\textbf{Consensus step}: Nodes run average consensus}:\\
              {
                %$v^i_{k-1} = \sum_{j\in Nb(i)}\pi_{ij}x^j_{k-1}$;
                $v^j_{k} = \sum_{l\in Nb(j)}\pi_{jl}x^l_{k}$;
              }\\
            \text{\textbf{Momentum step}}:\\
              {
                $\delta_k = \omega(x^j_k-x^j_{k-1})+(1-\omega)(v^j_k-v^j_{k-1})$;
              }\\
            \text{\textbf{Local gradient step} for node $j$}:\\
              {
                 $x^j_{k+1} = v^j_k - \alpha g^j(x^j_k) + \beta\delta_k$;
              }\\
            %   {\text{Option II}: $x^j_{k+1}=x^j_k-\alpha g^j(x^j_k)+\beta\delta_k$;
            %   }
          }
      }
\end{algorithm}

In the above update law, $g^j(x^j_k)$ is a stochastic gradient which is calculated by randomly selecting at uniform a mini-batch for each agent. Let $\mathcal{D}'$ be a mini-batch of the dataset $\mathcal{D}_j$ of the $j$-th agent. Therefore, $$g^j(x^j_k)=\frac{1}{b}\sum_{i\in\mathcal{D}'}\nabla f^i_j(x^j_k),$$ where $b$ is the size of $\mathcal{D}'$.

In~\cite{jiang2017collaborative,assran2018stochastic}, decentralized variants of classic momentum have been proposed (without analysis). On the other hand, our proposed DMSGD method uses a special parameter, $\omega$, to trade off between two different momentum terms. The first momentum term is implemented over the true \textit{decision} variables ($x_k^j$) while the second momentum term is implemented over the \textit{consensus} variables ($v_k^j$), which are graph-smoothed averages of the decision variables. 

We present a fairly general analysis for DMSGD; as special cases, we obtain known convergence properties for other methods. For example, we recover the decentralized classic momentum SGD by setting the parameter $\omega = 1$. When $\omega=0$, DMSGD produces a new decentralized MSGD algorithm in which the momentum relies on the consensus variables.
When the parameter $\beta$ is set to 0, the proposed DMSGD boils down to the decentralized SGD method without momentum~\cite{jiang2017collaborative,lian2017can}. Another slightly different alternative of DMSGD is to replace $v^j_k$ with $x^j_k$ in the local gradient step such that the consensus only affects the momentum term. The intuition behind this variant is that for the local gradient step, agent $j$ only relies on its current state information instead of the consensus, "refusing" to proceed the update on top of an "agreement".
For convenience and simplicity, the initial values of $x^j$ are set to 0 throughout the analysis. 

% \begin{rem}\label{adaptive}
% In our method, $\omega$ affects both convergence and consensus. As a heuristic, we also propose a variant where the value of $\omega$ varies based on the magnitude of the norms, $cm_1 =  \|x^j_k - x^j_{k-1}\|$ and $cm_2 =  \|v^j_k - v^j_{k-1}\|$. Based on the norm, the $\omega$ can be computed as $\omega = \frac{cm_1}{cm_1 + cm_2}$. This is inspired by methods that adaptively adjust the size of the gradient in momentum-based optimization, such as Adam~\cite{adam}.
% \end{rem}

We now rewrite the core update law with in a vector form as:\vspace{-10pt}
\begin{equation}
\begin{split}
    \label{rew_comp}
    \X_{k+1} &= \X_k - \alpha(\mathbf{g}(\X_k) + \frac{1}{\alpha}(I - \Pi)\X_k) \\
    &+ \beta(\omega I + (1-\omega)\Pi)(\X_k - \X_{k-1})
\end{split}
\end{equation}

Here, we define $\mathcal{S}(x_k) = \mathbf{g}(\mathbf{x}_k) + \frac{1}{\alpha}(I-\Pi)\mathbf{x}_k$ and $\tilde{\Pi} = \omega I + (1-\omega)\Pi$. Consequently, Eq.~\ref{rew_comp} can be written in a compact form as:
\vspace{-10pt}
\begin{equation}\label{compact1}
    \X_{k+1} = \X_k - \alpha \mathcal{S}(\X_k) + \beta \tilde{\Pi} (\X_k - \X_{k-1})
\end{equation}
% For option II, we can derive the analogous vector form such that\vspace{-10pt}
% \begin{equation}\label{compact2}
%     \X_{k+1} = \X_k - \alpha \mathbf{g}(\X_k) + \beta \tilde{\Pi} (\X_k - \X_{k-1})
% \end{equation}
The simplification in Eq.~\ref{compact1} enables us to construct a function that unifies the true objective function with a term that captures the constraint of consensus among agents (nodes of the communication graph).
\vspace{-10pt}
\begin{equation}\label{unified_objective_2}
    \mathcal{F}(\X) := F(\X) + \frac{1}{2\alpha}(\X^T(I-\Pi)\X)
\end{equation}
Comparing Eqs.~\ref{unified_objective_1} and~\ref{unified_objective_2}, we can know that they have exactly the same form and in our specific case corresponding to DMSGD, the parameter $\xi$ is the step size $\alpha$. 
% When $F$ is $\mu$-strongly convex, we immediately obtain that $\mathcal{F}(\x)$ is also strongly convex with parameter  $\mu'=\mu_m+\frac{1}{2\alpha}(1-\lambda_N)$, where $\lambda_N$ is the $N^{th}$-largest eigenvalue of $\Pi$.
$\mathcal{F}$ is smooth with $L'=L_m+\frac{1}{\alpha}(1-\lambda_2)$ where $\lambda_2$ is the second-largest eigenvalue of $\Pi$.
% Similarly, corresponding to Eq.~\ref{compact2}, one can obtain that the unified objective function is $F(\x)$ itself. 
% Throughout the rest of the analysis in the main paper, we only focus on the $\mathcal{F}(\x)$ (namely, DMSGD Option I) since all the convergence analysis techniques shown in the next section can directly apply to the case where the unified objective is $F(\x)$. For completeness, we present the specific analysis for $F(\x)$ in the supplementary materials. 
% For simplicity, in all the statements in the next section, we skip Option I.
\vspace{-10pt}
\section{Convergence Analysis}
\textbf{Consensus.} We first prove that the agents achieve consensus, i.e., each agent eventually obtains a parameter that is close to the ensemble average $\bar{x}_k = \frac{1}{N}\sum_{j=1}^Nx^j_k$, using the metrics of $\mathbb{E}[\|x^j_k-\bar{x}_k\|]$. In the setting of $d=1$, though $x^j_k$ and $\bar{x}_k$ are both scalars, we use the norm notation here for generality. As defined above, $\mathbf{x}$ has dimension of $N$. Define $\bar{\mathbf{x}}_k = [\bar{x}_k;\bar{x}_k;...;\bar{x}_k]_N$. Therefore, it holds that  $\|x^i_k-\bar{x}_k\|\leq\|\mathbf{x}_k-\bar{\mathbf{x}}_k\|$~\cite{berahas2018balancing} and instead of directly bounding $\|x^i_k-\bar{x}_k\|$, we investigate the upper bound for $\|\mathbf{x}_k-\bar{\mathbf{x}}_k\|$. We first obtain:

\begin{proposition}{(\textbf{Consensus})}\label{prop1}
Let all assumptions hold. The iterates generated by DMSGD satisfies the following inequality $\forall k \in \mathbb{N}$, $\exists\alpha> 0$:
\begin{equation}
    \mathbb{E}[||x^j_k - \bar{x}_k||] \leq \frac{8\alpha\sqrt{N}\sqrt{M^2 + \sigma^2}}{\sqrt{\eta(1-\beta\Lambda)}(1-\sqrt{\beta\Lambda})} ,
\end{equation}
where $\eta$ is defined as an arbitrarily small constant such that $\tilde{\Pi} \succcurlyeq \eta I$, $0<\eta <1$, $\Lambda = \omega + (1-\omega)\lambda_2$.
\end{proposition}
\begin{Proof}
The proof for this proposition is fairly technical and we provide the sketch here, referring interested readers to the supplementary materials. We first define $\tilde{\mathbf{x}}_k=\x_k-\bar{\x}_k$ and construct the linear time-invariant system for $[\tilde{\x}_{k+1};\tilde{\x}_k]$. Then by induction and setting initialization 0, we can express $[\tilde{\x}_{k+1};\tilde{\x}_k]$ using only the coefficient matrices and stochastic gradient inputs. By leveraging the decomposition techniques in matrices, the upper bound of matrix norms is obtained correspondingly. Hence, the iterates converge to the consensus estimate.
\end{Proof}
% For analysing the convergence rate for strongly convex functions, we state two lemmas.
\begin{rem}\label{rem:weakness}
Proposition~\ref{prop1} provides a uniform consensus error upper bound among agents, proportional to the step size $\alpha$ and the number of agents $N$ and inversely proportional to the gap between the largest and second-largest (in magnitude) eigenvalues of $\beta\tilde{\Pi}$. When $\omega=0$, DMSGD achieves the "best" consensus; the upper bound simplifies to $\frac{8\alpha\sqrt{N}\sqrt{M^2+\sigma^2}}{\sqrt{\eta(1-\beta\lambda_2)}(1-\sqrt{\beta\lambda_2})}$. When $\omega\to 1$, we get a worse-case upper bound on consensus error. Further, a more connected graph has a smaller value of $\lambda_2$, implying better consensus (which makes intuitive sense).
\end{rem}

\textbf{Nonconvex functions.} We summarize the main result on the convergence of DMSGD for nonconvex function in Theorem~\ref{thm1_main}. But first, we give an auxiliary lemma to simplify the proof process for Theorem~\ref{thm1_main}.
Throughout the rest of the analysis, $\mathcal{F}^*:=\mathcal{F}(\mathbf{x}^*)>-\infty$ is denoted as the minimum of the value sequence $\{\mathcal{F}(\x_k)\}, \forall k\in\mathbb{N}$. Recall the update law (Equation~\ref{compact1}). % as follows:
%\[\x_{k+1} =\x_k - \alpha \mathcal{S}(\x_k) + \beta\tilde{\Pi}(\mathbf{x}_k - \mathbf{x}_{k-1})\]
For convenience of analysis, we let $\mathbf{\hat{p}}_k = \beta\tilde{\Pi}(I-\beta\tilde{\Pi})^{-1}(\x_k-\x_{k-1})$ and rewrite the above equality in the following expression:
\begin{equation}\label{dmsgd_1}
\mathbf{x}_{k+1} + \mathbf{\hat{p}}_{k+1} = \mathbf{x}_{k} + \mathbf{\hat{p}}_{k} - \alpha(I-\beta\tilde{\Pi})^{-1}\mathcal{S}(\mathbf{x}_k)
\end{equation}
(Due to the space limit, we derive Eq.~\ref{dmsgd_1} in the supplementary materials.)
Let $\mathbf{\hat{z}}_k = \mathbf{x}_k + \mathbf{\hat{p}}_k$ such that the update rule finally becomes:
\begin{equation}\label{equiva}
\mathbf{\hat{z}}_{k+1} = \mathbf{\hat{z}}_{k} - \alpha(I-\beta\tilde{\Pi})^{-1}\mathcal{S}(\mathbf{x}_k)
\end{equation}
which resembles a regular form of SGD. Before showing the convergence analysis, we present a lemma for characterizing the main theorem.
\begin{lem}\label{lemma1_main}
Let all assumptions hold. The iterates $\{\mathbf{\hat{z}}_k\}$ generated by Eq.~\ref{equiva} satisfy: % the following inequality for all $k\in \mathbb{N}$
\begin{equation}
\mathbb{E}[\mathcal{F}(\mathbf{\hat{z}}_{k+1}) - \mathcal{F}(\mathbf{\hat{z}}_{k})]\leq -A_1\mathbb{E}[\|\nabla \mathcal{F}(\mathbf{x}_k)\|^2] + A_2
\end{equation}
where $A_1 =\frac{\alpha}{2(1-\beta\Lambda)}-\frac{L'\alpha^2}{2(1-\beta\Lambda)^2}, A_2 =\frac{\alpha^3 L'^2(\beta\Lambda)^2}{(1-\beta\Lambda)^5}(M^2 + \sigma^2) + \frac{L'\alpha^2\sigma^2}{2(1-\beta\Lambda)^2}$.
\end{lem}
The proof is presented in the supplementary materials. With the above lemma in hand, we are ready to state the main theorem for the nonconvex analysis for DMSGD. Before that, we discuss the choice of step size for the convergence analysis. While constant step size enables algorithms to converge faster, diminishing step size can achieve better accuracy in stochastic optimization. In this context, the step size should satisfy a condition that can guarantee the value sequence $\{\mathcal{F}(\mathbf{x})\}$ to sufficiently descend. According to Lemma~\ref{lemma1_main}, $\alpha$ should satisfy $\alpha\leq\frac{1-\beta\Lambda}{L'}$.
\begin{theo}\label{thm1_main}
Let all assumptions hold. Suppose the step size satisfies $\alpha=\textnormal{min}\{\frac{1-\beta\Lambda}{2L'}, \sqrt{\frac{N}{K}}\}$. The iterates $\{\mathbf{x}_k\}$ generated by Eq.~\ref{compact1} satisfy the following inequality:
\begin{equation}
\begin{split}
&\frac{1}{K}\sum^K_{k=1}\mathbb{E}[\|\nabla \mathcal{F}(\mathbf{x}_k)\|^2]\leq \textnormal{max}\bigg\{\frac{8(\mathcal{F}(\mathbf{x}_1)-\mathcal{F}^*)L'}{K},\\& \frac{4(1-\beta\Lambda)(\mathcal{F}(\mathbf{x}_1)-\mathcal{F}^*)}{\sqrt{NK}}\bigg\}
+\frac{4NL'^2(\beta\Lambda)^2(M^2+\sigma^2)}{K(1-\beta\Lambda)^4}\\&+\frac{2NL'\sigma^2}{(1-\beta\Lambda)\sqrt{NK}} .
\end{split}
\end{equation}
\end{theo}
\begin{Proof}
Using the conclusion from Lemma~\ref{lemma1_main}, by induction, we can get the desired results. Please refer to the supplementary materials for more details.
\end{Proof}
% We can observe that the step size in Theorem~\ref{thm1_main} is complex when the variance is existing and the positiveness needs verification. In the supplementary materials, we show that $\alpha>0$ naturally.  

Theorem~\ref{thm1_main} shows that with a properly selected constant stepsize, for nonconvex functions, DMSGD can converge to the optimal solution $\x^*$ (which essentially is a stationary point) with a rate of $\mathcal{O}(1/K+1/\sqrt{NK})$. This matches the best results in~\cite{wang2019slowmo,yu2019linear}. Additionally, the selection of $\alpha$ satisfies the condition that $\alpha\leq\frac{1-\beta\Lambda}{L'}$ such that when $K\geq\frac{NL'^2}{(1-\beta\Lambda)^2}$, Theorem~\ref{thm1_main} suggests $\mathcal{O}(1/\sqrt{NK})$, which implies the linear speed up for DMSGD. Additional analytical results regarding strongly convex and quasi-convex objective functions are presented in the supplementary materials due to the space limit.
%\end{rem}

%% file: NewAnalysisSupp.tex
\subsection{Proof for Consensus}
We denote by $\Lambda$ the equivalent second large eigenvalue of $\tilde{\Pi}$ such that $\Lambda = \omega+(1-\omega)\lambda_2$. $\lambda_2< 1$ and $\omega\in[0,1)$ such that $\Lambda< 1$ for the analysis. Though $\omega$ can be set 1, in practice, DMSGD requires it to be strictively less than 1. 

\textbf{Proposition~\ref{prop1}}: 
Let all assumptions hold, the iterates generated by DMSGD satisfies the following inequality $\forall k \in \mathbb{N}$, $\exists\alpha> 0$,
\begin{equation}
    \mathbb{E}[||x^j_k - \bar{x}_k||] \leq \frac{8\alpha\sqrt{N}\sqrt{M^2 + \sigma^2}}{\sqrt{\eta(1-\beta\Lambda)}(1-\sqrt{\beta\Lambda})}
\end{equation}
where, $\eta$ is defined as an arbitrarily small positive constant such that $\tilde{\Pi} \succcurlyeq \eta I$, $\eta <1$, $\Lambda = \omega + (1-\omega)\lambda_2$.

\begin{Proof}
We define
\[\tilde{x}^j_k = x^j_k-\bar{x}_k\]
the compact form of which is written as
\[\tilde{\X}_k=\x_k-\bar{\x}_k\].

% the compact form of which is written as
% \[\tilde{\X}_k=\x_k-\bar{\x}_k\]
Similarly, we have $\tilde{\X}_{k+1} = \X_{k+1}-\bar{\x}_{k+1}$ and next construct the linear time-invariant system for $[\tilde{\x}_{k+1};\tilde{\X}_k]$. Let $J = \frac{1}{N}\mathbf{1}\mathbf{1}^T$, where $\mathbf{1}$ is a $N\times 1$ dimension vector with entries being 1. Substituting the update law~\ref{compact1} into $\tilde{\X}_{k+1}$, we have
\begin{align}
    \tilde{\X}_{k+1} =&  (I-J)\X_{k} - \alpha(I-J)\mathcal{S}(\X_k)\label{prop1_eqn1} \\
                        & + \beta(I-J)\tilde{\Pi}(\X_k - \X_{k-1})\nonumber\\
    \text{As}\;(I-J)\tilde{\Pi} & = (I-J)(\omega I + (1-\omega)\Pi)\nonumber\\
     & = \omega (I-J) + (1-\omega)(I-J)\Pi\nonumber\\
     &=  \omega (I-J) + (1-\omega)\Pi(I-J)\nonumber\\
     &=  \omega I (I-J) + (1-\omega)\Pi(I-J)\nonumber\\
     &=  [\omega I + (1-\omega)\Pi](I-J)\nonumber\\
     &=  \tilde{\Pi}(I-J)\nonumber
\end{align}
Where the third inequality holds due to $\Pi$ being doubly stochastic, then Eq.~\ref{prop1_eqn1} can be rewritten as
\begin{align*}
    \tilde{\X}_{k+1} & = \tilde{\X}_k - \alpha(I-J)\mathcal{S}(\X_k) + \beta\tilde{\Pi}(I-J)(\X_k - \X_{k-1})\\
        & =  \tilde{\X}_k - \alpha(I-J)\mathcal{S}(\X_k) + \beta\tilde{\Pi}(\tilde{\X}_k - \tilde{\X}_{k-1})\\
\end{align*}
Hence, we can obtain the linear time-invariant system for $[\tilde{\X}_{k+1}; \tilde{\X}_k]$.
\begin{equation}\label{eq25}
\begin{aligned}
\begin{bmatrix}
\tilde{\x}_{k+1}\\
\tilde{\x}_{k}\end{bmatrix}&=\begin{bmatrix}
I +\beta\tilde{\Pi} & -\beta\tilde{\Pi}\\
I & 0
\end{bmatrix}\\&\begin{bmatrix}
\tilde{\x}_k\\
\tilde{\x}_{k-1}
\end{bmatrix}-\alpha\begin{bmatrix}
(I-J)\mathcal{S}(\X_k)\\
0
\end{bmatrix}
\end{aligned}
\end{equation}
Eq.~\ref{eq25} shows an equivalent linear time-invariant system with an input with respect to stochastic gradient.

We define $\tilde{X}_{k+1} = \begin{bmatrix}
\tilde{\X}_{k+1}\\
\tilde{\X}_{k}
\end{bmatrix}$,\\ 
$ \mathbf{A} = \begin{bmatrix}
I+\beta\tilde{\Pi} & -\beta\tilde{\Pi}\\
I & 0
\end{bmatrix} ,\; \mathbf{B}_k = \begin{bmatrix}
(I-J)\mathcal{S}(x_k)\\
0
\end{bmatrix} $ such that\\
\begin{align}\label{eq26}
    \tilde{X}_{k+1} &= \mathbf{A}\tilde{X}_k - \alpha \mathbf{B}_k
\end{align}
We perform induction on $k$ for Eq.~\ref{eq26} and have the following relationship.
\begin{align*}
    \tilde{X}_{k+1} &= \mathbf{A}^k \tilde{X}_0 - \alpha\sum_{s=0}^k \mathbf{A}^{k-s} \mathbf{B}_s
\end{align*}
The next step of the proof is to analyze $\mathbf{A}^{k-s},\;\forall k\geq s$ using Schur decomposition and to bound the norm of it. It can directly follow from the proof of Theorem~4 in \cite{jakovetic2014fast} and we can arrive at 
\begin{align*}
    || \mathbf{A}^{k-s}|| \leq \frac{8(\sqrt{\beta\Lambda})^{k-s}}{\sqrt{\eta(1-\beta\Lambda)}}.
\end{align*}
We next bound the norm of $\mathbf{B}_k, \;\forall k \geq 0$. $\mathbf{B}_k$ only consists of $(I-J)\mathcal{S}(\x_k)$ and $0$ such that essentially it is a column vector in this context. Therefore, we have
\begin{align*}
    ||\mathbf{B}_k|| = \lvert\lvert\begin{bmatrix}
    (I-J)\mathcal{S}(x_k)\\
    0
    \end{bmatrix}\rvert\rvert \leq ||I-J||\,||\mathcal{S}(\X_k)||\sqrt{N},
\end{align*}
By taking the expectation on both sides, we have
\begin{align*}
    \mathbb{E}[||\mathbf{B}_k||] \leq \mathbb{E}[||\mathcal{S}(\X_k)||]\sqrt{N} \leq \sqrt{M^2 + \sigma^2}\sqrt{N},
\end{align*}
Hence, with initialization being $0$, we have
\begin{align*}
    ||\tilde{X}_k|| = \alpha||\sum^k_{s=0}\mathbf{A}^{k-s}\mathbf{B}_s|| \leq \alpha \sum_{s=0}^k ||\mathbf{A}^{k-s}|| ||\mathbf{B}_s||
\end{align*}
Taking expectation on both sides yields
\begin{align*}
    \mathbb{E}[||\tilde{X}||] &\leq \alpha\sum_{s=0}^k ||\mathbf{A}^{k-s}|| \mathbb{E}[||\mathbf{B}_s||]\\
    &\leq\alpha\sum^k_{s=0}\frac{8(\sqrt{\beta\Lambda})^{k-s}}{\sqrt{\eta(1-\beta\Lambda})}\sqrt{N}\sqrt{M^2+\sigma^2}\\
    &\leq \frac{8\alpha\sqrt{N}\sqrt{M+\sigma^2}}{\sqrt{\eta(1-\beta\Lambda)}(1-\sqrt{\beta\Lambda})}
\end{align*}
which completes the proof with $||x^j_k - \bar{x}_k|| \leq ||\tilde{\X}_k|| \leq ||\tilde{X}_k||$.
\end{Proof}
% *********************************
% \textbf{Contents from the main draft}
\subsection{Derivation for Eq.~\ref{dmsgd_1}}
We derive the update law Eq.~\ref{dmsgd_1}.  
Recalling Eq.~\ref{compact1}, we have
\[\x_{k+1}=\x_{k}-\alpha\mathcal{S}(\x_k)+\beta\tilde{\Pi}(\x_k-\x_{k-1})\]
Let $\mathbf{\hat{p}}_k = \beta\tilde{\Pi}(I-\beta\tilde{\Pi})^{-1}(\x_k-\x_{k-1})$ such that $\mathbf{\hat{p}}_{k+1} = \beta\tilde{\Pi}(I-\beta\tilde{\Pi})^{-1}(\x_{k+1}-\x_k)$. According to this, we have
\begin{equation}
\begin{split}&\x_{k+1}+\mathbf{\hat{p}}_{k+1}=\x_{k+1}+\beta\tilde{\Pi}(I-\beta\tilde{\Pi})^{-1}(\x_{k+1}-\x_k)\\
&=(I+\beta\tilde{\Pi}(I-\beta\tilde{\Pi})^{-1})\x_{k+1}-\beta\tilde{\Pi}(I-\beta\tilde{\Pi})^{-1}\x_k\\
&=(I+\beta\tilde{\Pi}(I-\beta\tilde{\Pi})^{-1})(\x_{k}-\alpha\mathcal{S}(\x_k)+\beta\tilde{\Pi}(\x_k-\x_{k-1}))\\
&-\beta\tilde{\Pi}(I-\beta\tilde{\Pi})^{-1}\x_k\\
&=\x_k - \alpha(I_{Nd}+\beta\tilde{\Pi}(I-\beta\tilde{\Pi})^{-1})\mathcal{S}(\x_k)+(I\\
&+\beta\tilde{\Pi}(I-\beta\tilde{\Pi})^{-1})\beta\tilde{\Pi}(\x_k-\x_{k-1})
\end{split}
\end{equation}
As $I+\beta\tilde{\Pi}(I-\beta\tilde{\Pi})^{-1}=(I-\beta\tilde{\Pi})(I-\beta\tilde{\Pi})^{-1}+\beta\tilde{\Pi}(I-\beta\tilde{\Pi})^{-1}=(I-\beta\tilde{\Pi})^{-1}$, we have
\begin{equation}
\begin{split}
\x_{k+1}+\mathbf{\hat{p}}_{k+1}&=\x_k-\alpha(I-\beta\tilde{\Pi})^{-1}\mathcal{S}(\x_k)\\&+(I-\beta\tilde{\Pi})^{-1}\beta\tilde{\Pi}(\x_k-\x_{k-1})
\end{split}
\end{equation}
Since $\beta\tilde{\Pi}(I-\beta\tilde{\Pi})^{-1}=\beta\tilde{\Pi}\sum_{l=0}^{\infty}(\beta\tilde{\Pi})^l=\sum_{l=0}^{\infty}(\beta\tilde{\Pi})^l\beta\tilde{\Pi}=(I-\beta\tilde{\Pi})^{-1}\beta\tilde{\Pi}$, it completes the derivation for the update law Eq.~\ref{dmsgd_1}, i.e., $\x_{k+1}+\mathbf{\hat{p}}_{k+1}=\x_{k}+\mathbf{\hat{p}}_{k}-\alpha(I-\beta\tilde{\Pi})^{-1}\mathcal{S}(\x_k)$.
\subsection{Proof for Lemma~\ref{lemma1_main} and Theorem~\ref{thm1_main}}
\textbf{Lemma}~\ref{lemma1_main}
Let all assumptions hold. The iterates $\{\mathbf{\hat{z}}_k\}$ generated by Eq.~\ref{equiva} satisfy: % the following inequality for all $k\in \mathbb{N}$
\begin{equation}
\mathbb{E}[\mathcal{F}(\mathbf{\hat{z}}_{k+1}) - \mathcal{F}(\mathbf{\hat{z}}_{k})]\leq -A_1\mathbb{E}[\|\nabla \mathcal{F}(\mathbf{x}_k)\|^2] + A_2
\end{equation}
where $A_1 =\frac{\alpha}{2(1-\beta\Lambda)}-\frac{L'\alpha^2}{2(1-\beta\Lambda)^2}, A_2 =\frac{\alpha^3 L'^2(\beta\Lambda)^2}{(1-\beta\Lambda)^5}(M^2 + \sigma^2) + \frac{L'\alpha^2\sigma^2}{2(1-\beta\Lambda)^2}$.
\begin{proof}
Based on Definition~\ref{smooth}, the following relationship can be obtained
\begin{equation}
\begin{split}
&\mathcal{F}(\mathbf{\hat{z}}_{k+1}) - \mathcal{F}(\mathbf{\hat{z}}_{k})\leq\\
&-\alpha\nabla \mathcal{F}(\mathbf{\hat{z}}_k)^\top(I-\beta\tilde{\Pi})^{-1}\mathcal{S}(\mathbf{x}_k) + \frac{L'\alpha^2}{2}\|(I-\beta\tilde{\Pi})^{-1}\mathcal{S}(\mathbf{x}_k)\|^2\\
&=-\alpha [(I-\beta\tilde{\Pi})^{-\frac{1}{2}}\nabla \mathcal{F}(\mathbf{\hat{z}}_k)]^\top(I-\beta\tilde{\Pi})^{-\frac{1}{2}}\mathcal{S}(\mathbf{x}_k) \\
&+ \frac{L'\alpha^2}{2}\|(I-\beta\tilde{\Pi})^{-1}\mathcal{S}(\mathbf{x}_k)\|^2
\end{split}
\end{equation}
Taking the expectation on the both sides results in:
\begin{equation}
\begin{split}
&\mathbb{E}[\mathcal{F}(\mathbf{\hat{z}}_{k+1}) - \mathcal{F}(\mathbf{\hat{z}}_{k})]\\
&\leq-\alpha \mathbb{E}[((I-\beta\tilde{\Pi})^{-\frac{1}{2}}\nabla \mathcal{F}(\mathbf{\hat{z}}_k))^\top(I-\beta\tilde{\Pi})^{-\frac{1}{2}}\nabla\mathcal{F}(\mathbf{x}_k)] \\&+ \frac{L'\alpha^2}{2(1-\beta\Lambda)^2}\sigma^2+\frac{L'\alpha^2}{2}\mathbb{E}[\|(I-\beta\tilde{\Pi})^{-\frac{1}{2}}\\&(I-\beta\tilde{\Pi})^{-\frac{1}{2}}\nabla\mathcal{F}(\mathbf{x}_k)\|^2]\\
&\leq-\alpha\mathbb{E}[((I-\beta\tilde{\Pi})^{-\frac{1}{2}}\nabla \mathcal{F}(\mathbf{\hat{z}}_k))^\top(I-\beta\tilde{\Pi})^{-\frac{1}{2}}\nabla\mathcal{F}(\mathbf{x}_k)] \\&+ \frac{L'\alpha^2}{2(1-\beta\Lambda)^2}\sigma^2+\frac{L'\alpha^2}{2(1-\beta\Lambda)}\mathbb{E}[\|(I-\beta\tilde{\Pi})^{-\frac{1}{2}}\nabla\mathcal{F}(\mathbf{x}_k)\|^2]\\&+\frac{\alpha}{2}\mathbb{E}[\|(I-\beta\tilde{\Pi})^{-\frac{1}{2}}\nabla\mathcal{F}(\mathbf{x}_k)\|^2]\\
&-\frac{\alpha}{2}\mathbb{E}[\|(I-\beta\tilde{\Pi})^{-\frac{1}{2}}\nabla\mathcal{F}(\mathbf{x}_k)\|^2] \\&+ \frac{\alpha}{2}\mathbb{E}[\|(I-\beta\tilde{\Pi})^{-\frac{1}{2}}\nabla\mathcal{F}(\mathbf{\hat{z}}_k)\|^2]
\end{split}
\end{equation}
Therefore, the last inequality can be cast as
\begin{equation}\label{eq39}
\begin{split}
&\mathbb{E}[\mathcal{F}(\mathbf{\hat{z}}_{k+1}) - \mathcal{F}(\mathbf{\hat{z}}_{k})]\\
&\leq\frac{\alpha}{2}\mathbb{E}[\|(I-\beta\tilde{\Pi})^{-\frac{1}{2}}\nabla\mathcal{F}(\mathbf{\hat{z}}_k) - (I-\beta\tilde{\Pi})^{-\frac{1}{2}}\nabla\mathcal{F}(\mathbf{x}_k)\|^2]\\& + \frac{L'\alpha^2\sigma^2}{2(1-\beta\Lambda)^2}\\
&+\bigg[\frac{L'\alpha^2}{2(1-\beta\Lambda)}-\frac{\alpha}{2}\bigg]\mathbb{E}[\|(I-\beta\tilde{\Pi})^{-\frac{1}{2}}\nabla\mathcal{F}(\mathbf{x}_k)\|^2]
\end{split}
\end{equation}
It can observed that from the right hand side of the last inequality, the first term can be bounded above based on the definition of smoothness. Therefore, we have the following relationship
\begin{equation}
\begin{split}
&\mathbb{E}[\|(I-\beta\tilde{\Pi})^{-\frac{1}{2}}\nabla\mathcal{F}(\mathbf{\hat{z}}_k) - (I-\beta\tilde{\Pi})^{-\frac{1}{2}}\nabla\mathcal{F}(\mathbf{x}_k)\|^2]\\
&\leq\frac{1}{1-\beta\Lambda}\mathbb{E}[\|\nabla\mathcal{F}(\mathbf{\hat{z}}_k)-\nabla\mathcal{F}(\mathbf{x}_k)\|^2]\\
&\leq\frac{1}{1-\beta\Lambda} L'^2\mathbb{E}[\|\mathbf{\hat{z}}_k - \mathbf{x}_k\|^2]=\frac{1}{1-\beta\Lambda}L'^2\mathbb{E}[\|\mathbf{\hat{p}}_k\|^2]\\
&=\frac{1}{1-\beta\Lambda}L'^2\mathbb{E}[\|\beta\tilde{\Pi}(I-\beta\tilde{\Pi})^{-1}(\mathbf{x}_k-\mathbf{x}_{k-1})\|^2]\\
&\leq\frac{L'^2(\beta\Lambda)^2}{(1-\beta\Lambda)^3}\mathbb{E}[\|\mathbf{x}_k - \mathbf{x}_{k-1}\|^2]
\end{split}
\end{equation}
The last inequality follows from the property of submultiplicative norm and the following relationship
\begin{equation}
\begin{split}
&\|(I - \beta\tilde{\Pi})^{-1}\|=\|\sum_{t=0}^\infty(\beta\tilde{\Pi})^t\|=\|\sum_{t=0}^\infty\beta^t\tilde{\Pi}^t\|\leq\sum_{t=0}^\infty\beta^t\|\tilde{\Pi}\|^t\\
&\leq\sum_{t=0}^\infty\beta^t\Lambda^t=\frac{1}{1-\beta\Lambda}
\end{split}
\end{equation}
The last equality holds as $\beta< 1$ and $\Lambda< 1$. Based on the update law~\ref{compact1} and use induction, we can obtain that $\mathbf{x}_k - \mathbf{x}_{k-1} = -\alpha\sum_{t=0}^{k-1}(\beta\tilde{\Pi})^t\mathcal{S}(\mathbf{x}_{k-1-t})$. Substituting such an equality into the last inequality, with the result from Lemma~\ref{lemma2} presented below, yields
\begin{equation}\label{eq42}
\begin{split}
&\mathbb{E}[\|(I-\beta\tilde{\Pi})^{-\frac{1}{2}}\nabla\mathcal{F}(\mathbf{\hat{z}}_k) - (I-\beta\tilde{\Pi})^{-\frac{1}{2}}\nabla\mathcal{F}(\mathbf{x}_k)\|^2]\\
&\leq\frac{L'^2\alpha^2(\beta\Lambda)^2(M^2+\sigma^2)}{(1-\beta\Lambda)^5}
\end{split}
\end{equation}
Hence, combining Eq.~\ref{eq39} and Eq.~\ref{eq42} we have
\begin{equation}
\begin{split}
&\mathbb{E}[\mathcal{F}(\mathbf{\hat{z}}_{k+1}) - \mathcal{F}(\mathbf{\hat{z}}_{k})]\leq\frac{L'^2\alpha^3(\beta\Lambda)^2(M^2+\sigma^2)}{(1-\beta\Lambda)^5} + \frac{L'\alpha^2\sigma^2}{2(1-\beta\Lambda)^2}\\
&+\bigg[\frac{L'\alpha^2}{2(1-\beta\Lambda)^2}-\frac{\alpha}{2(1-\beta\Lambda)}\bigg]\mathbb{E}[\|\nabla\mathcal{F}(\mathbf{x}_k)\|^2]
\end{split}
\end{equation}
Changing the sign before the third term on the right hand side of the last inequality leads to the desired results.
\end{proof}
\textbf{Theorem}~\ref{thm1_main}
Let all assumptions hold. Suppose the step size satisfies $\alpha=\textnormal{min}\{\frac{1-\beta\Lambda}{2L'}, \sqrt{\frac{N}{K}}\}$. The iterates $\{\mathbf{x}_k\}$ generated by Eq.~\ref{compact1} satisfy the following inequality:
\begin{equation}
\begin{split}
&\frac{1}{K}\sum^K_{k=1}\mathbb{E}[\|\nabla \mathcal{F}(\mathbf{x}_k)\|^2]\leq \textnormal{max}\bigg\{\frac{8(\mathcal{F}(\mathbf{x}_1)-\mathcal{F}^*)L'}{K},\\& \frac{4(1-\beta\Lambda)(\mathcal{F}(\mathbf{x}_1)-\mathcal{F}^*)}{\sqrt{NK}}\bigg\}
+\frac{4NL'^2(\beta\Lambda)^2(M^2+\sigma^2)}{K(1-\beta\Lambda)^4}\\&+\frac{2NL'\sigma^2}{(1-\beta\Lambda)\sqrt{NK}}.
\end{split}
\end{equation}
\begin{Proof}
Recalling the conclusion of Lemma~\ref{lemma1_main}, we have
\[\mathbb{E}[\mathcal{F}(\mathbf{\hat{z}}_{k+1}) - \mathcal{F}(\mathbf{\hat{z}}_{k})]\leq -A_1\mathbb{E}[\|\nabla \mathcal{F}(\mathbf{x}_k)\|^2] + A_2\]
By induction, the following relationship can be obtained:
\begin{equation}
A_1\sum_{k=1}^{K}\mathbb{E}[\|\nabla \mathcal{F}(\mathbf{x}_k)\|^2]\leq\mathbb{E}[\mathcal{F}(\mathbf{\hat{z}}_1)-\mathcal{F}(\mathbf{\hat{z}}_K)]+KA_2,
\end{equation}
which suggests
% As $(m+1)\textnormal{min}_{k=0,1,...,m}\mathbb{E}[\|\nabla \mathcal{C}(\mathbf{x}_k)\|^2]\leq\sum_{k=0}^{m}\mathbb{E}[\|\nabla \mathcal{C}(\mathbf{x}_k)\|^2]$,
% the following inequality can be obtained:
\begin{equation}
\sum_{k=1}^K\mathbb{E}[\|\nabla \mathcal{F}(\mathbf{x}_k)\|^2]\leq\frac{\mathcal{F}(\mathbf{\hat{z}}_1)-\mathcal{F}(\mathbf{\hat{z}}_K)}{A_1} + \frac{KA_2}{A_1}
\end{equation}
Since the minimum value is denoted as $\mathcal{F}^*$. Therefore, the last inequality becomes as follows
\begin{equation}
\frac{1}{K}\sum_{k=1}^K\mathbb{E}[\|\nabla \mathcal{F}(\mathbf{x}_k)\|^2]\leq\frac{\mathcal{F}(\mathbf{\hat{z}}_1)-\mathcal{F}^*}{KA_1} + \frac{A_2}{A_1}
\end{equation}
As the step size satisfies $\alpha = \textnormal{min}\{\frac{1-\beta\Lambda}{2L'}, \sqrt{\frac{N}{K}}\}$, then we have $\frac{\alpha}{4(1-\beta\Lambda)}\leq A_1$, which leads the last inequality to
\begin{equation}
\begin{split}
\frac{1}{K}\sum_{k=1}^K\mathbb{E}[\|\nabla \mathcal{F}(\mathbf{x}_k)\|^2]&\leq\frac{4(1-\beta\Lambda)(\mathcal{F}(\mathbf{\hat{z}}_1)-\mathcal{F}^*)}{K\alpha} \\&+ \frac{4(1-\beta\Lambda)A_2}{\alpha}
\end{split}
\end{equation}
Substituting $A_2$, the step size $\alpha$, $\mathbf{\hat{z}}_1=\mathbf{x}_1$ into the last inequality yields
\begin{equation}
\begin{split}
&\frac{1}{K}\sum^K_{k=1}\mathbb{E}[\|\nabla \mathcal{F}(\mathbf{x}_k)\|^2]\leq \textnormal{max}\bigg\{\frac{8(\mathcal{F}(\mathbf{x}_1)-\mathcal{F}^*)L'}{K},\\& \frac{4(1-\beta\Lambda)(\mathcal{F}(\mathbf{x}_1)-\mathcal{F}^*)}{\sqrt{NK}}\bigg\}
+\frac{4NL'^2(\beta\Lambda)^2(M^2+\sigma^2)}{K(1-\beta\Lambda)^4}\\&+\frac{2NL'\sigma^2}{(1-\beta\Lambda)\sqrt{NK}},
\end{split}
\end{equation}
which is the desired result. The second and third terms on the right hand side of the last inequality is due to $\alpha\leq \sqrt{\frac{N}{K}}$.
\end{Proof}
\subsection{Analysis for Strongly Convex and Quasi-convex Functions}
\textbf{Strongly convex functions.} Next, we study the convergence for strongly convex objective functions; in this case, the algorithm converges to a globally optimal solution. We show convergence by upper bounding $\mathbb{E}[\mathcal{F}(\x)-\mathcal{F}^*]$. Before showing the main results, we present the definition of strong convexity and two auxiliary lemmas.
\begin{defn}\label{strong_convexity}
A function $f:\mathbb{R}^d\to \mathbb{R}$ is $\mu$-strongly convex, if for all $x,y\in \mathbb{R}^d$, we have
$$f(y)\geq f(x)+\nabla f(x)^T(y-x)+\frac{\mu}{2}\|y-x\|^2.$$
\end{defn}
When the local objective functions $f_j$ are $\mu_j$-strongly convex, then $\sum_{j=1}^{N}f_j(x^j)$ is $\mu_m$-strongly convex where $\mu_m:=\textnormal{min}\{\mu_1, \mu_2, ..., \mu_N\}$.
We then immediately obtain that $\mathcal{F}(\x)$ is also strongly convex with parameter  $\mu'=\mu_m+\frac{1}{2\alpha}(1-\lambda_N)$, where $\lambda_N$ is the $N^{th}$-largest eigenvalue of $\Pi$.
\begin{lem}\label{lemma1}
Let all stated assumptions hold. Suppose that $\mathcal{F}(\x)$ for all $\x \in \mathbb{R}^d$ is $\mu'$-strongly convex. Then the following relationship holds:\vspace{-10pt}
\begin{equation}
    -\langle \nabla \mathcal{F}(\X), \nabla\mathcal{F}(\X) \rangle \leq -\mu'^2||\X - \X^*||^2 \leq -\frac{2\mu'^2}{L'}(\mathcal{F}(\X) - \mathcal{F^*})
\end{equation}
where $\X^* = \argmin_{\X\in\mathbb{R}^N} {\mathcal{F}(\X)}$,\; $\mathcal{F^*}:= \mathcal{F}(\mathbf{x}^*) > -\infty$.
\end{lem}
\begin{Proof}
As $\mathcal{F}(\X)$ is $\mu'$-strongly convex, we can immediately obtain the following relationship based on its definition,
\begin{equation}
    (\nabla \mathcal{F}(\x)-\nabla\mathcal{F}(\mathbf{y}))^T(\x-\mathbf{y})\geq\mu'\|\x-\mathbf{y}\|^2
\end{equation}
for all $\x,\mathbf{y}\in\mathbb{R}^d$. Using Cauthy-Schwarz inequality, we have
\begin{equation}
    \|\nabla \mathcal{F}(\x)-\nabla\mathcal{F}(\mathbf{y})\|\|\x-\mathbf{y}\|\geq\mu'\|\x-\mathbf{y}\|^2
\end{equation}
Let $\mathbf{y}=\x^*$ such that
\begin{equation}
    ||\nabla \mathcal{F}(\X) - \nabla \mathcal{F}(\x^*)|| \geq \mu' || \X - \X^*||
\end{equation}
Since $\nabla \mathcal{F^*} = 0$, we have 
\begin{equation}
    -\langle \nabla \mathcal{F}(\X), \nabla\mathcal{F}(\X) \rangle \leq -\mu'^2||\X - \X^*||^2 
\end{equation}
Further, based on the definition of smoothness, we have 
\begin{equation}
    \mathcal{F}(\X) \leq \mathcal{F}^* + \frac{L'}{2}||\X - \X^*||^2
\end{equation}
combining the last two inequalities completes the proof.
\end{Proof}
\begin{lem}\label{lemma2}
Let all assumptions hold. Then the iterates generated by DMSGD $\forall k \in \mathbb{N}$, $\exists\alpha> 0$ satisfy the following relationship:

\begin{equation}
    \mathbb{E}[||\X_{k+1} - \X_k||^2] \leq \frac{\alpha^2(M^2 + \sigma^2)}{(1-\beta\Lambda)^2}
\end{equation}
where $\Lambda = \omega + (1-\omega)\lambda_2$. Alternatively, we have a tighter bound as
\begin{equation}
    \mathbb{E}[||\X_{k+1} - \X_k||^2] \leq \frac{(1-(\Lambda\beta)^{k+1})^2\alpha^2(M^2 + \sigma^2)}{(1-\Lambda\beta)^2}
\end{equation}
\end{lem}
\begin{Proof}
Recall the update law in a vector form:
\[    \X_{k+1} = \X_k - \alpha \mathcal{S}(\X_k) + \beta \tilde{\Pi} (\X_k - \X_{k-1})\] such that
\begin{equation}
    \X_{k+1} - \X_k = - \alpha \mathcal{S}(\X_k) + \beta \tilde{\Pi} (\X_k - \X_{k-1})
\end{equation}
By induction, we can have
\begin{align}
    \X_{k+1} - \X_k &= - \alpha \sum^k_{m=0} \beta^{k-m} \tilde{\Pi}^{k-m} \mathcal{S}(\X_m)\nonumber
\end{align}
Let $\mathcal{T}_{k+1} = \sum^k_{m=0}(\beta\Lambda)^m = \frac{1-(\beta\Lambda)^{k+1}}{1-\beta\Lambda}$, then
\begin{align}
    ||\X_{k+1} - \X_k||^2 &= \alpha^2 \lvert\lvert \sum^k_{m=0} \frac{\beta^{k-m}}{\mathcal{T}_{k+1}} \tilde{\Pi}^{k-m} \mathcal{S}(\X_m)\rvert\rvert^2 \mathcal{T}^2_{k+1}\nonumber\\
    &\leq \alpha^2\mathcal{T}^2_{k+1}\sum^k_{m=0} \frac{\beta^{k-m}||\tilde{\Pi}||^{k-m}}{\mathcal{T}_{k+1}}\lvert\lvert\mathcal{S}(\X_m)\rvert\rvert^2\nonumber\\
    &\leq \alpha^2\mathcal{T}_{k+1}\sum^k_{m=0} (\beta\Lambda)^{k-m}\lvert\lvert\mathcal{S}(\X_m)\rvert\rvert^2
\end{align}
Here, the second inequality follows from the convexity of $||\cdot||^2$ and Jensen's inequality. The last inequality follows from the property of doubly stochastic matrix. Taking the expectation on both sides results in 
\begin{align}
    \mathbb{E}[||\X_{k+1} - \X_k||^2] &\leq \frac{(1-(\Lambda\beta)^{k+1})^2\alpha^2(M^2 + \sigma^2)}{(1-\Lambda\beta)^2}\nonumber\\
    &\leq \frac{\alpha^2(M^2 + \sigma^2)}{(1-\Lambda\beta)^2}, \nonumber
\end{align}
which completes the proof.
\end{Proof}
We now state the main result for convergence under strongly convex functions.
\begin{theo}{(\textbf{Strongly convex case})}\label{thm1}
Let all assumptions hold. Suppose that $\mathcal{F}$ is $\mu'$-strongly convex. The iterates $\{\X_k\}$ generated by DMSGD with $0<\alpha\leq \frac{L'}{2\mu'^2}$ and $0\leq \beta < 1$ satisfy the following inequality for all $k \in \mathbb{N}$:

\begin{align*}
    \mathbb{E}[\mathcal{F}(\X_{k+1}) - \mathcal{F}^*]& \leq (1 - \frac{2\alpha\mu'^2}{L'})\mathbb{E}[\mathcal{F}(\X_{k}) - \mathcal{F}^*] + \alpha M\sigma\\& + \alpha M \frac{\sqrt{M^2 + \sigma^2}}{1-\beta\Lambda} + \frac{L'\alpha^2(M^2 + \sigma^2)}{2(1-\beta\Lambda)^2},
\end{align*}
where $\Lambda = \omega + (1-\omega)\lambda_2$. Alternatively, we can have a tighter bound as follows:
\begin{align*}
    \mathbb{E}[\mathcal{F}(\X_{k+1}) - \mathcal{F}^*] \leq& (1 - \frac{2\alpha\mu'^2}{L'})\mathbb{E}[\mathcal{F}(\X_{k}) - \mathcal{F}^*] + \alpha M\sigma\\& + \alpha M \sqrt{M^2 + \sigma^2} \frac{1-(\beta\Lambda)^k}{1-\beta\Lambda} \\&+ \frac{L'(1-(\beta\Lambda)^{k+1})^2\alpha^2(M^2 + \sigma^2)}{2(1-\beta\Lambda)^2}.
\end{align*}
\end{theo}
\begin{Proof}
Based on the definition of smoothness, we can obtain that
\begin{align*}
    \mathcal{F}_{k+1} - \mathcal{F}_k & \leq \langle \nabla \mathcal{F}(\X_k), \X_{k+1} - \X_k \rangle + \frac{L'}{2}||\X_{k+1} - \X_k||^2\\
    &= \langle \nabla \mathcal{F}(\X_k), -\alpha\mathcal{S}(\X_k) + \beta\tilde{\Pi}(\X_k - \X_{k-1}) \rangle\\
    & + \frac{L'}{2}||\X_{k+1} - \X||^2\\
    & = \langle \nabla \mathcal{F}(\X_k), -\alpha \nabla \mathcal{F}(\X_k) \rangle \\
    & + \langle \nabla \mathcal{F}(\X_k), \alpha \nabla \mathcal{F}(\X_k) -\alpha\mathcal{S}(\X_k) \rangle\\
    & +  \langle \nabla \mathcal{F}(\X_k), \beta\tilde{\Pi}(\X_k - \X_{k-1}) \rangle\\
    & + \frac{L'}{2}||\X_{k+1} - \X_k||^2
\end{align*}
According to Lemma~\ref{lemma1}, we have
\begin{align*}
    \mathcal{F}(\X_{k+1}) - \mathcal{F}(\X_{k}) &\leq \frac{-2\alpha\mu'^2}{L'}(\mathcal{F}(\X_k) - \mathcal{F}^*)\\
    & + \alpha||\nabla \mathcal{F}(\X_k)||||\nabla \mathcal{F}(\X_k) - \mathcal{S}(\X_k)||\\
    & + \alpha||\nabla \mathcal{F}(\X_k)||\\&\sum^{k-1}_{m=0}\beta^{k-1-m}||\tilde{\Pi}^{k-1-m}\mathcal{S}(\X_m)||\\
    & + \frac{L'}{2}||\X_{k+1} - \X||^2
\end{align*}
The last inequality follows from Cauchy-Schwarz inequality. Hence, we have
\begin{align*}
    \mathcal{F}(\X_{k+1}) - \mathcal{F}^* &\leq \mathcal{F}(\X_{k}) - \mathcal{F}^* - \frac{2\alpha\mu'^2}{L'}(\mathcal{F}(\X_k) - \mathcal{F}^*)\\
    & + \alpha||\nabla \mathcal{F}(\X_k)||||\nabla \mathcal{F}(\X_k) - \mathcal{S}(\X_k)||\\
    & + \alpha||\nabla \mathcal{F}(\X_k)||\sum^{k-1}_{m=0}(\beta\Lambda)^{k-1-m}||\mathcal{S}(\X_m)||\\
    & + \frac{L'}{2}||\X_{k+1} - \X||^2\\
\end{align*}

Rearranging the last inequality and taking expectations on both sides yields
\begin{align*}
    \mathbb{E}[\mathcal{F}(\X_{k+1}) - \mathcal{F}^*] &\leq (1- \frac{2\alpha\mu'^2}{L'})\mathbb{E}[\mathcal{F}(\X_k) - \mathcal{F}^*]\\
    & + \alpha||\nabla\mathcal{F}(\X_k)||\mathbb{E}[||\nabla \mathcal{F}(\X_k) - \mathcal{S}(\X_k)||]\\
    & + \alpha||\nabla \mathcal{F}(\X_k)||\\&\sum^{k-1}_{m=0}(\beta\Lambda)^{k-1-m}\mathbb{E}[||\mathcal{S}(\X_m)||]\\
    & + \frac{L'}{2}\mathbb{E}[||\X_{k+1} - \X||^2]
\end{align*}

\begin{align*}
    \text{As }\; \mathbb{E}[||\nabla \mathcal{F}(\X_k) - \mathcal{S}(\X_k)||] &= \sqrt{(\mathbb{E}[||\nabla \mathcal{F}(\X_k) - \mathcal{S}(\X_k)||])^2}\\
    &\leq \sqrt{\mathbb{E}[||\nabla \mathcal{F}(\X_k) - \mathcal{S}(\X_k)||^2]}\\
    &= \sqrt{\sigma^2}\\
    &= \sigma
\end{align*}

Substituting the bound of $\mathcal{S}(\X_k)$ and using Lemma~\ref{lemma2}, we have
\begin{align*}
    &\mathbb{E}[||\nabla \mathcal{F}(\X_k) - \mathcal{S}(\X_k)||] \leq (1- \frac{2\alpha\mu'^2}{L'})\mathbb{E}[\mathcal{F}(\X_k) - \mathcal{F}^*]\\
    & + \alpha M \sigma + \alpha M \sqrt{M^2 + \sigma^2}\frac{1-(\beta\Lambda)^k}{1-\beta\Lambda}\\
    & + \frac{L'\alpha^2(M^2 + \sigma^2)(1-(\beta\Lambda)^{k+1})^2}{2(1-\beta\Lambda)^2}\\
    &\leq (1- \frac{2\alpha\mu'^2}{L'})\mathbb{E}[\mathcal{F}(\X_k) - \mathcal{F}^*] + \alpha M \sigma + \alpha M \frac{\sqrt{M^2 + \sigma^2}}{1-\beta\Lambda} \\&+ \frac{L'\alpha^2(M^2 + \sigma^2)}{2(1-\beta\Lambda)^2},
\end{align*}
which completes the proof.
\end{Proof}
Immediately, Theorem~\ref{thm1} implies the following corollary to show that with a properly set constant stepsize $\alpha$, DMSGD enables the iterates $\{\X_k\}$ to converge to a solution that is proportional to $\alpha$ in a linear rate.
\begin{corollary}\label{coro1}
Let all stated assumptions hold. The iterates generated by DMSGD with $0<\alpha<\frac{L'}{2\mu'^2}$ and $0\leq\beta<1$ satisfy the following relationship $\forall k \in \mathbb{N}$:\vspace{-8pt}
\begin{align*}
    \mathbb{E}[\mathcal{F}(\X_k) - \mathcal{F}^*] &\leq \frac{RL'}{2\alpha\mu'^2}\\& + (1-\frac{2\alpha\mu'^2}{L'})^{k-1}(\mathcal{F}(\X_1) - \mathcal{F}^* - \frac{RL'}{2\alpha\mu'^2}),
\end{align*}
where $R = \alpha M\sigma + \alpha M \sqrt{M^2 + \sigma^2}\frac{1}{1-\beta\Lambda} + \frac{L'\alpha^2(M^2 + \sigma^2)}{2(1-\beta\Lambda)^2}$.
\end{corollary}
\begin{Proof}
Using Theorem~\ref{thm1}, we can obtain that
\begin{align*}
    \mathbb{E}[\mathcal{F}(\X_k) - \mathcal{F}^*] \leq (1- \frac{2\alpha\mu'^2}{L'})\mathbb{E}[\mathcal{F}(\X_k) - \mathcal{F}^*] + R,\\
\end{align*}
subtracting the constant $\frac{RL'}{2\alpha\mu'^2}$ from both sides, one can get
\begin{align*}
    \mathbb{E}[\mathcal{F}(\X_k) - \mathcal{F}^*] - \frac{RL'}{2\alpha\mu'^2} &\leq (1- \frac{2\alpha\mu'^2}{L'})\mathbb{E}[\mathcal{F}(\X_k) - \mathcal{F}^*]\\& + R - \frac{RL'}{2\alpha\mu'^2}\\
    &= (1- \frac{2\alpha\mu'^2}{L'})(\mathbb{E}[\mathcal{F}(\X_k) - \mathcal{F}^*] \\ &- \frac{RL'}{2\alpha\mu'^2})\\
\end{align*}
We can know that the above inequality is a contraction inequality since
$0<\frac{2\alpha\mu'^2}{L'}\leq1$ due to $0<\alpha\leq\frac{L'}{2\mu'^2}$. Hence the desired result is obtained by applying the inequality repeatedly through iteration $k\in\mathbb{N}$.
\end{Proof}
%\begin{Proof}
%The proof is easily obtained by repeatedly using Theorem~\ref{thm1} along with time $k$.
%\end{Proof}
%\begin{rem}
Corollary~\ref{coro1} suggests that when $k\to\infty$, DMSGD enables the iterates $\{\x_k\}$ to converge within $\frac{RL'}{2\alpha\mu^{'2}}$ from the optimal point $\x^*$. The error bound is essentially with respect to the variance of stochastic gradient and the network error among agents when substituting $R$ into $\frac{RL'}{2\alpha\mu^{'2}}$. Also, when $\sigma=0$ and $\alpha\to 0$, the iterates $\{\x_k\}$ converges to $\x^*$. 
%\end{rem}
%\begin{rem}

Due to the fact that $\mathbb{E}[F(\x_k)]\leq\mathbb{E}[\mathcal{F}(\x_k)]$ and that $F(\x^*)=\mathcal{F}(\x^*)$, the sequence of true objective function values are bounded above by $\mathbb{E}[F(\x_k)-F(\x^*)]\leq\mathbb{E}[\mathcal{F}(\x_k)-\mathcal{F}(\x^*)]$. Hence, Corollary~\ref{coro1} also implies that we can establish the analogous convergence rates in terms of the true objective function value sequence $\{F(\x_k)\}$. 
%\end{rem}

\textbf{Polyak-\L{}ojasiewicz Inequality}. It is well known that in the deterministic case, under gradient descent with strongly-convex and smoothness assumptions, a global linear convergence rate can be achieved. Recently, SGD and other stochastic variants have also shown to have linear convergence~\cite{karimi2016linear,gower2019sgd}.  However, most models, even simple deep learning models, are not strongly-convex. In this context, we introduce the Polyak-\L{}ojasiewicz (PL) inequality, which enables the gradient to grow as a quadratic function of sub-optimality. With PL condition and smoothness, using gradient descent, it is shown to achieve a linear rate. Further, in ~\cite{karimi2016linear,gower2019sgd}, with constant step size, the stochastic gradient descent can converge to a solution that is proportional to $\alpha$ in a linear rate. The important property of PL condition is that it does not imply convexity (hence called as quasi-convexity~\cite{gower2019sgd} or invexity~\cite{karimi2016linear}). Therefore, with the motivation to relax the strongly-convex assumption, we show that with PL condition, using a constant stepsize, DMSGD can achieve linear convergence rate to converge to the neighborhood of $\X^*$. To the best of our knowledge, it is the first time to show that PL condition enables a decentralized momentum SGD to converge at a linear rate.
\vspace{-5pt}
\begin{theo}{(\textbf{Quasi convex case})}\label{thm2}
Let all assumptions hold. Suppose that $\mathcal{F}(\X)$ satisfies the Polyak-\L{}ojasiewicz inequality such that
\begin{align*}
    ||\nabla\mathcal{F}(\X)||^2 \geq 2\hat{\mu}(\mathcal{F}(\X) - \mathcal{F}^*) ,\; \forall \X \in \mathbb{R}^N
\end{align*}
The iterates $\{\X_k\}$ generated by DMSGD with $0 < \alpha \leq \frac{1}{2\hat{\mu}}$ and $0\leq\beta<1$ satisfy the following inequality for all $k\in\mathbb{N}$
\begin{align*}
    \mathbb{E}[\mathcal{F}(\X_k) - \mathcal{F}^*] \leq &\frac{R}{2\alpha\hat{\mu}^2} + (1-2\alpha\hat{\mu})^{k-1}(\mathcal{F}(\X_1) - \mathcal{F}^*\\ &- \frac{R}{2\alpha\hat{\mu}^2}),
\end{align*}
where $R = \alpha M\sigma + \alpha M \sqrt{M^2 + \sigma^2}\frac{1}{1-\beta\Lambda} + \frac{L'\alpha^2(M^2 + \sigma^2)}{2(1-\beta\Lambda)^2}$.
\end{theo}
\begin{Proof}
According to the definition of smoothness, we have
\begin{align*}
    \mathcal{F}(\X_{k+1}) - \mathcal{F}(\X_k) & \leq \langle \nabla \mathcal{F}(\X_k), \X_{k+1} - \X_k \rangle\\
            & + \frac{L'}{2}||\X_{k+1} - \X_k||^2\\
    & = \langle \nabla \mathcal{F}(\X_k), - \alpha \mathcal{S}(\X_k)+\beta\tilde{\Pi}(\X_k-\X_{k-1}) \rangle \\
    & + \frac{L'}{2}||\X_{k+1} - \X_k||^2\\
    & = \langle \nabla \mathcal{F}(\X_k), - \alpha\nabla \mathcal{F}(\X_k) \rangle \\
    & + \langle \nabla \mathcal{F}(\X_k), \alpha\nabla \mathcal{F}(\X_k)- \alpha \mathcal{S}(\X_k) \rangle \\
    & + \langle \nabla \mathcal{F}(\X_k), \beta\tilde{\Pi}(\X_k-\X_{k-1}) \rangle \\
    & + \frac{L'}{2}||\X_{k+1} - \X_k||^2
\end{align*}
As $|| \nabla \mathcal{F}(\X_k)||^2 \geq 2\hat{\mu}(\mathcal{F}(\X_k)-\mathcal{F}^*)$, we get
\begin{align*}
    -\alpha|| \nabla \mathcal{F}(\X_k)||^2\leq-2\alpha\hat{\mu}(\mathcal{F}(\X_k)-\mathcal{F}^*)
\end{align*}
We skip the rest of the proof as it directly follows from the proof of Theorem~\ref{thm1} and Corollary~\ref{coro1} immediately.
\end{Proof}
%\begin{rem}
It can be observed that with only smoothness and PL condition, DMSGD still converges in a linear rate to the neighborhood. However, it should be noted that the strong convexity can imply PL condition, but not vice versa. That is the reason why we use $\hat{\mu}$, not $\mu'$. Since when objective function $\mathcal{F}(\X)$ is strongly convex, $-|| \nabla \mathcal{F}(\X_k)||^2\leq -2\mu'(\mathcal{F}(\X_k)-\mathcal{F}^*)$ holds. 